\pdfoutput=1

\documentclass[11pt]{article}

\usepackage[]{EMNLP2022}

\usepackage{times}
\usepackage{latexsym}

\usepackage[T1]{fontenc}

\usepackage[utf8]{inputenc}

\usepackage{microtype}

\usepackage{inconsolata}

\usepackage[ruled,vlined,linesnumbered]{algorithm2e}
\usepackage{amsmath}
\usepackage{amssymb}
\usepackage{booktabs}
\usepackage{microtype}
\usepackage[capitalize]{cleveref}
\usepackage{multirow,multicol}
\usepackage{algpseudocode}
\usepackage{graphics}
\usepackage{graphicx}

\title{SAT: Improving Semi-Supervised Text Classification with Simple Instance-Adaptive Self-Training}

\author{Hui Chen \quad Wei Han \quad Soujanya Poria\\
  \large{Singapore University of Technology and Design}\\
  \texttt{\{hui\_chen, wei\_han\}@mymail.sutd.edu.sg}\\
  \texttt{sporia@sutd.edu.sg}\\
  }

\begin{document}
\maketitle
\begin{abstract}
Self-training methods have been explored in recent years and have exhibited great performance in improving semi-supervised learning. This work presents a \textbf{S}imple instance-\textbf{A}daptive self-\textbf{T}raining method (SAT) for semi-supervised text classification. SAT first generates two augmented views for each unlabeled data and then trains a meta-learner to automatically identify the relative strength of augmentations based on the similarity between the original view and the augmented views. The weakly-augmented view is fed to the model to produce a pseudo-label and the strongly-augmented view is used to train the model to predict the same pseudo-label. We conducted extensive experiments and analyses on three text classification datasets and found that with varying sizes of labeled training data, SAT consistently shows competitive performance compared to existing semi-supervised learning methods. Our code can be found at \url{https://github.com/declare-lab/SAT.git}.
\end{abstract}

\section{Introduction}
Pretrained language models have achieved extremely good performance in a wide range of natural language understanding tasks~\cite{devlin2019bert}. However, such performance often has a strong dependence on large-scale high-quality supervision. Since labeled linguistic data needs large amounts of time, money, and expertise to obtain, improving models' performance in few-shot scenarios (i.e., there are only a few training examples per class) has become a challenging research topic. 

Semi-supervised learning in NLP has received increasing attention in improving performance in few-shot scenarios, where both labeled data and unlabeled data are utilized~\cite{berthelot2019mixmatch,sohn2020fixmatch,li2021semi}. Recently, several self-training methods have been explored to obtain task-specific information in unlabeled data. UDA~\cite{xie2020unsupervised} applied data augmentations to unlabeled data and proposed an unsupervised consistency loss that minimizes the divergence between different unlabeled augmented views. To give self-training more supervision, MixText~\cite{chen2020mixtext,berthelot2019mixmatch} employed Mixup~\cite{zhang2018mixup,chen2022doublemix} to learn an intermediate representation of labeled and unlabeled data. Both UDA and MixText utilized consistency regularization and confirmed that such regularization exhibits outstanding performance in semi-supervised learning. To simplify the consistency regularization process, FixMatch~\cite{sohn2020fixmatch} classified two unlabeled augmented views into a weak view and a strong view, and then minimized the divergence between the probability distribution of the strong view and the pseudo label of the weak view. However, in NLP, it is hard to distinguish the relative strength of augmented text by observation, and randomly assigning an augmentation strength will limit the performance of FixMatch on text.

To tackle this problem in FixMatch, our paper introduces an instance-adaptive self-training method SAT, where we propose two criteria based on a classifier and a scorer to automatically identify the relative strength of augmentations on text. Our main contributions are:
\begin{itemize}
    \item First, we apply popular data augmentation techniques to generate different views of unlabeled data and design two novel criteria to calculate the similarity between the original view and the augmented view of unlabeled data in FixMatch, boosting its performance on text.
    \item We then conduct empirical experiments and analyses on three few-shot text classification datasets. Experimental results confirm the efficacy of our SAT method.
\end{itemize}
 

\section{Method}

\subsection{Problem Setting}
\label{sec:problem_setting}
In this work, we learn a model to map an input $x\in \mathcal{X}$ onto a label $y\in \mathcal{Y}$ in text classification tasks. In semi-supervised learning, we use both labeled examples and unlabeled examples during training. Let $\mathcal{X} = \{(x_b, y_b): b \in (1,...,B)\}$ be a batch of $B$ labeled examples, where $x_b$ are the training examples and $y_b$ are labels. Let $\mathcal{U} = \{u_b: b \in (1,...,\mu B)  \}$ be a batch of $\mu B$ unlabeled examples, where $\mu$ is a hyperparameter which determines the relative sizes of $\mathcal{X}$ and $\mathcal{U}$.

\subsection{SAT}
The entire process of SAT is illustrated in~\cref{alg:sat}. Similar to common semi-supervised learning methods, our approach consists of a supervised part and an unsupervised part. Our supervised part minimizes the cross-entropy loss between the labeled data and their targets. Our unsupervised part first generates two unlabeled augmented views, then applies an augmentation choice network to determine the relative augmentation strength, and finally calculates a consistency loss between the probability distribution of the strongly-augmented view and the pseudo label of the weakly-augmented view. Since the relative augmentation strength in our SAT method has no direct correlation to the augmentation techniques, our semi-supervised learning process can be more adaptive to the training data, compared to FixMatch.

The augmentation choice network is trained by the labeled data and we design two criteria to train it where (1) one is based on a \textbf{classifier} and (2) the other is based on a \textbf{scorer}. \cref{alg:line2} to \cref{alg:line7} in~\cref{alg:sat} shows how we train the augmentation choice network. For each labeled data, we first calculate the similarity between the original data and its augmented variants, respectively, and then rank the augmented samples according to the similarity scores. In our classifier-based criterion, we employ a \textbf{cross-entropy loss} to measure the distance, while in our scorer-based criterion, we calculate the \textbf{cosine similarity}. Afterward, we define the one with a higher similarity score as the weakly-augmented sample and use it to train the augmentation choice network. For our classifier-based method, we apply a \textbf{cross-entropy loss} as the training objective. For our scorer-based method, we use a \textbf{contrastive loss}~\cite{chen2020simple} to update the network. Finally, the trained augmentation choice network is used to automatically identify the augmentation strength in unlabeled data.

\begin{algorithm}[ht!]
\SetAlgoLined
\KwIn{$\mathcal{D}^{train} = \{\mathcal{X}, \mathcal{U}\}$ where $\mathcal{X}=\{(x_b,y_b):b\in(1,...,B)\}$ and $\mathcal{U}=\{u_b: b\in(1,...,\mu B)\}$; 
augmentation methods $\alpha_1$, $\alpha_2$; 
main network $f(;\theta)$ with parameters $\theta$ and its probability distribution $p$; augmentation choice network $G(;\theta_G)$ with parameters $\theta_G$;
criteria $\mathcal{C}$, $\Gamma$; cross-entropy loss $H$; 
unlabeled loss weight $\lambda_u$; confidence threshold $\tau$; learning rates $\beta, \eta$}
\KwOut{Updated network weights $\theta$}

\tcp{Calculate supervised loss}
$\mathit{l}_s = \frac{1}{B}\sum_{b=1}^B H(y_b, p(y|x_b))$

\For{$(x_b,y_b) \in \mathcal{X}$}{
\label{alg:line2}
    $i_1^b,i_2^b = \mathcal{C}(\alpha_1(x_b),x_b,y_b),\mathcal{C}(\alpha_2(x_b),x_b,y_b)$ \\
    $i_w^b, i_s^b = \mathrm{Descending}(i_1^b, i_2^b)$\\
}

\tcp{Update the augmentation choice network}
$\mathit{l}_{aug\_choice} =\frac{1}{B}\sum_{b=1}^B \Gamma(x_b, \alpha_1(x_b), \alpha_2(x_b), i_w^b)$ \\
$\theta_G = \theta_G - \beta \nabla \mathit{l}_{aug\_choice}$ \\
\label{alg:line7}

\For{each $u_b\in \mathcal{U}$}{
    $\hat{i}_w^b, \hat{i}_s^b = G(u_b, \alpha_1(u_b), \alpha_2(u_b);\theta_G)$
}
\tcp{Calculate unsupervised loss}
$\mathit{l}_u=\frac{1}{\mu B}\sum_{b=1}^{\mu B}\mathbf{1}\{\mathrm{max}(p(y|\alpha_{\hat{i}_w^b}(u_b)))>\tau\} H(\mathrm{argmax}(p(y|\alpha_{\hat{i}_w^b}(u_b))), p(y|\alpha_{\hat{i}_s^b}(u_b)))$

\tcp{Total loss: add up supervised loss and unsupervised loss}
$\mathit{l}_{total} = \mathit{l}_s + \lambda_u \mathit{l}_u$ \\
\tcp{Update the main network}
$\theta = \theta - \eta \nabla l_{total}$
\caption{SAT: Simple Instance-Adaptive Self-Training}
\label{alg:sat}
\end{algorithm}

\begin{table*}[ht]
    \centering
    \scalebox{0.82}{
    \begin{tabular}{l|cccccccc}
    \toprule
    \multirow{2}{*}{\textbf{Methods}} & \multicolumn{2}{c}{\textbf{AG News} ($c=4$)} & \multicolumn{2}{c}{\textbf{Yahoo!} ($c=10$)} & \multicolumn{2}{c}{\textbf{IMDB} ($c=2$)} & & \\
    & Acc.(\%) & F1.(\%) & Acc.(\%) & F1.(\%) & Acc.(\%) & F1.(\%) & Average & $\Delta$\\
    \midrule
    BERT~\cite{devlin2019bert} & $69.18_{3.7}$ & $68.27_{3.5}$ & $58.11_{1.6}$ & $57.38_{1.9}$ & $63.16_{1.4}$ & $62.93_{1.6}$ & $63.17$&-\\
    UDA~\cite{xie2020unsupervised} & $76.69_{3.2}$ & $76.51_{3.0}$ & $59.32_{2.0}$ & $58.47_{2.3}$ & $64.88_{1.7}$ & $64.57_{1.5}$ & $66.74$&$+3.57$\\
    MixText~\cite{chen2020mixtext} & $78.07_{2.8}$ & $77.23_{3.5}$ & $59.93_{1.9}$ & $59.24_{1.8}$ & $65.22_{1.1}$ & $65.78_{1.2}$ & $67.58$&$+4.41$\\
    FixMatch~\cite{sohn2020fixmatch} & $80.22_{2.4}$ & $79.98_{2.1}$ & $60.17_{1.7}$ & $59.86_{1.5}$ & $64.52_{1.6}$ & $64.31_{1.4}$ & $68.18$&$+5.01$\\
    \midrule
    SAT: classifier-based (Ours) & $\mathbf{86.38_{2.0}}$ & $\mathbf{86.29_{2.3}}$ & $\mathbf{61.51_{1.8}}$ & $\mathbf{61.09_{1.6}}$ & $65.43_{1.2}$ & $64.28_{1.4}$ & $70.83$&$+7.66$\\
    SAT: scorer-based (Ours) & $85.43_{1.2}$ & $85.30_{1.5}$ & $61.33_{1.5}$ & $60.96_{1.4}$ & $\mathbf{68.96_{1.7}}$ & $\mathbf{68.92_{1.6}}$ & $\mathbf{71.82}$&$\mathbf{+8.65}$\\
    \bottomrule
    \end{tabular}}
    \caption{Accuracy (\%) and Macro F1 (\%) on three diverse text classification tasks for BERT, UDA, MixText, FixMatch, and our SAT method. $c$: number of classes; $\Delta$: improvement compared with BERT.}
    \label{tab:main_results}
\end{table*}

\section{Experimental Setup}
We conducted empirical experiments to compare our approach with a couple of existing semi-supervised learning methods on a variety of text classification benchmark datasets.
\subsection{Datasets and Metrics}
We considered three diverse few-shot text classification scenarios in our experiments: AG News which categorizes more than 1 million news articles into 4 categories --- World, Sports, Business, and Sci/Tech~\cite{zhang2015character}, and Yahoo! Answers which classifies question-answer pairs into 10 clusters where all question-answer pairs in a cluster ask about the same thing~\cite{zhang2015character}, and IMDB which predicts sentiment of movie reviews to be positive or negative~\cite{maas2011learning}.

We used the original test set as our test set and randomly sampled from the training set to construct the training unlabeled set and development set. To balance the class distribution in the experiments, we randomly sampled $N_c$ samples per class to be used for training. In AG News and IMDB, $N_c = 10$. In the Yahoo! dataset, we set $N_c$ as 20 to ensure consistent results. For all experiments, we used accuracy (\%) and macro F1 score (\%) as the evaluation metrics.

\subsection{Baselines}
To test the effectiveness of our approach, we compared it with several popular self-training methods: UDA~\cite{xie2020unsupervised}, MixText~\cite{chen2020mixtext}, FixMatch~\cite{sohn2020fixmatch}. To ensure fair comparisons, we used the same augmentation techniques\footnote{The implementation is based on \url{https://github.com/makcedward/nlpaug}.}, i.e., back-translation~\cite{sennrich2016improving} and synonym replacement~\cite{wei2019eda}, in all baselines. For back-translation augmentation, we used German as the middle language. For synonym replacement augmentation, the substitution percentage is 30\%. Also, we used the same BERT-based-uncased model, unlabeled data size, and batch size in all methods.

\section{Results}

\subsection{Main Results}
This section compares our SAT method with BERT, UDA, MixText, and FixMatch on three text classification datasets. Our main results are shown in~\cref{tab:main_results}. Results are averaged over five different runs.

We observed that BERT achieves a mean score across our three datasets of 63.17\%, and UDA improves performance noticeably by
+3.57\%. MixText and FixMatch show better performance in improving the BERT baseline, where the mean score increases are +4.41\% and +5.01\%, respectively. Our classifier-based SAT method, which applies a cross-entropy loss to measure the similarity between the original data and its augmented variants, achieves a mean score of 70.83\%, outperforming FixMatch by +2.65\%. The scorer-based SAT, which employs cosine distance to measure the similarity further improves about 1\% over the classifier-based SAT.

From these results, we obtained the following findings. Firstly, strategically selecting the strongly and weakly augmented samples in self-training can effectively boost performance. Compared to FixMatch, we improved the performance by adding a lightweight meta-learner to automatically identify augmentation strengths, without sacrificing much training time. Secondly, when tackling datasets with few training examples, using cosine distance to measure the similarity between two examples and using contrastive loss to train the meta-learner shows better and more robust performance.

\subsection{Ablation: Size of Labeled Data}
\begin{figure}[ht]
    \centering
    \includegraphics[width=0.96\columnwidth]{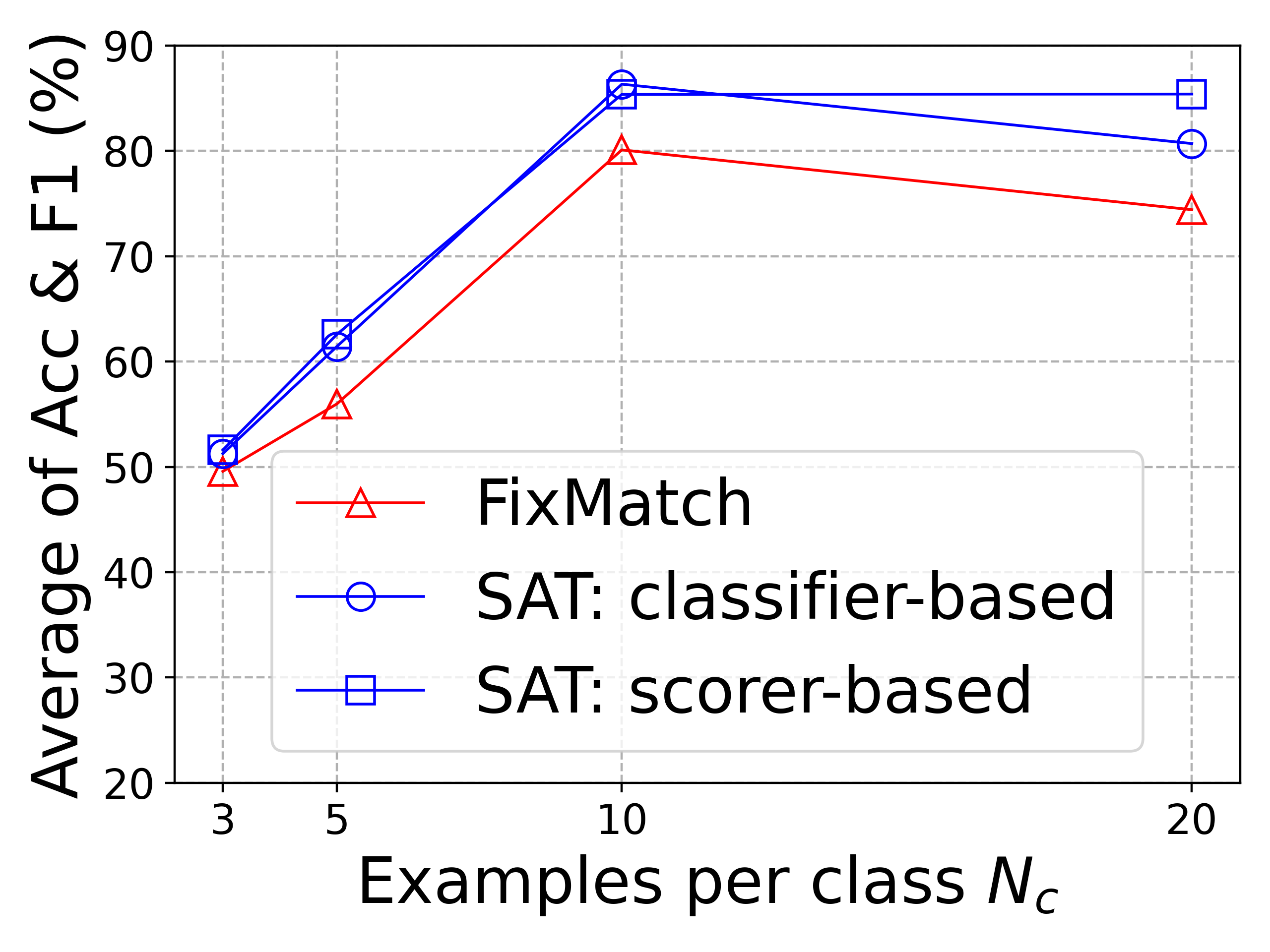}
    \caption{Average scores of accuracy and macro F1 from FixMatch and our method for different sizes of labeled data on the AG News dataset.}
    \label{fig:ablation_data_size}
\end{figure}

This ablation investigates how our semi-supervised learning method performs for different sizes of labeled data. \cref{fig:ablation_data_size} compares the average scores of accuracy and macro F1 of FixMatch and our methods. First, we observed that our methods can consistently outperform FixMatch with varying data sizes. This indicates that strategically selecting the strongly and weakly augmented samples contributes to the final performance in self-training. Second, when $N_c$ increases from 3 to 10, the scores of the three methods increase accordingly. When $N_c$ becomes 20, the performance of FixMatch and the classifier-based SAT drops, which is consistent with prior findings on the diminished effect of data augmentation for larger datasets~\cite{xie2020unsupervised,andreas2020good}. However, the scorer-based SAT does not show an obvious performance decrease, showing that in few-shot datasets, the scorer-based method is more robust than the classifier-based method.

\subsection{Ablation: Various Augmentation Techniques}
\label{sec:aug_methods}
To evaluate the effect of various augmentation techniques in our method, we performed experiments using different text augmentation technique combinations in our method. These augmentation techniques are widely-used: \textbf{(1) Synonym Replacement (SR)} substitutes words with WordNet synonyms~\cite{wei2019eda}; \textbf{(2) Pervasive Dropout (PD)} applies a word-level dropout with a probability of 0.1 on text~\cite{sennrich2016edinburgh}; \textbf{(3) Random Insertion (RI)} randomly inserts words in a sentence~\cite{wei2019eda}; \textbf{(4) Back-translation (BT)} translates text into another language and then back into the original language~\cite{sennrich2016improving}.

\begin{figure}[ht]
    \centering
    \includegraphics[width=0.96\columnwidth]{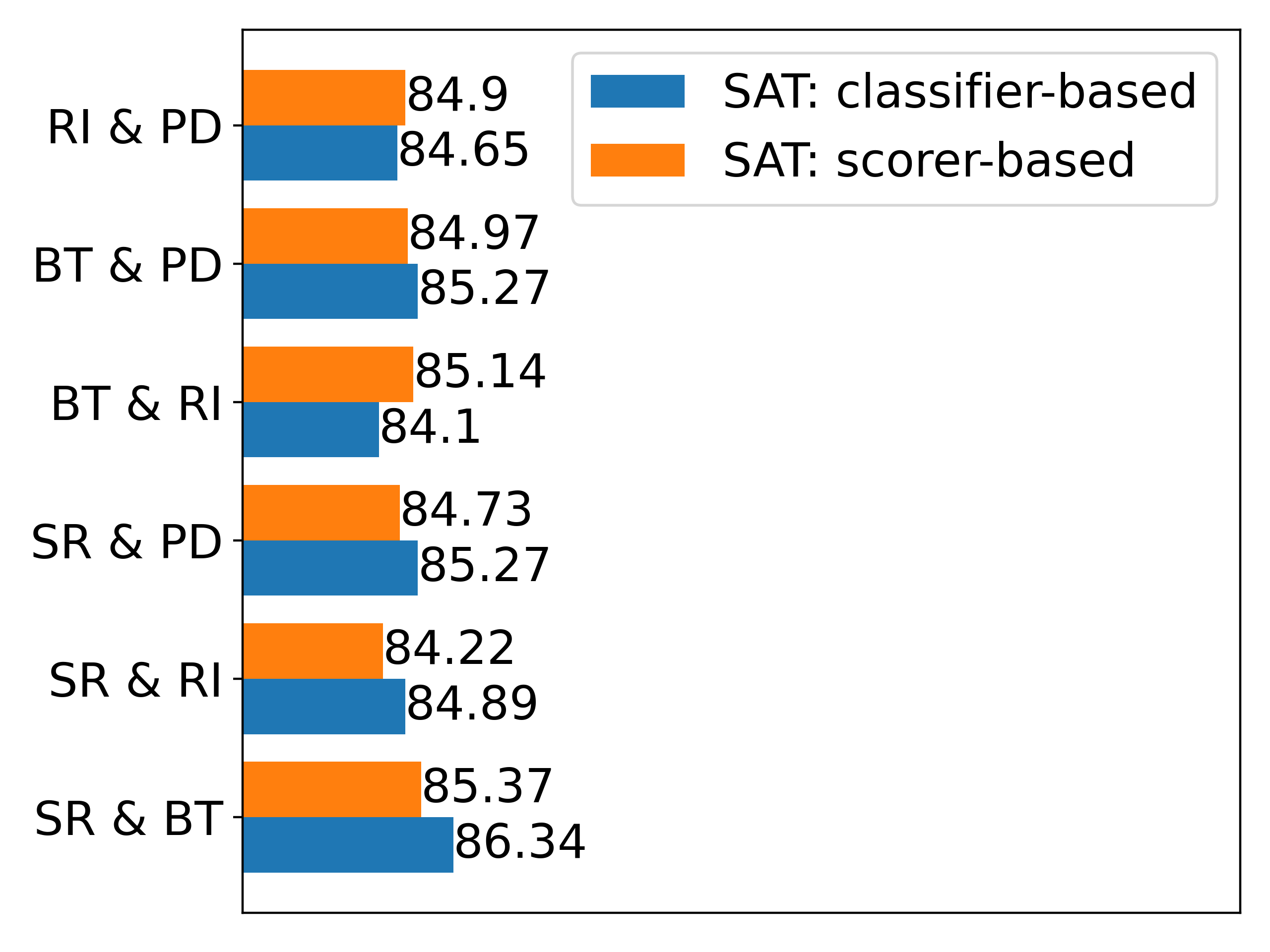}
    \caption{Average scores of accuracy and macro F1 from different augmentation technique combinations in our method on the AG News dataset, where $N_c=10$.}
    \label{fig:ablation_aug_methods}
\end{figure}

\cref{fig:ablation_aug_methods} compares average scores of accuracy and macro F1 from different augmentation techniques in our method on the AG News dataset. The combination of back-translation and synonym replacement improves performance the best, perhaps because they maintain a good balance between injecting proper perturbation noise and preserving the original meaning of the text.

\section{Related Work}
Our work combines data augmentation and consistency regularization to improve semi-supervised learning and is inspired by prior work in these areas. Recently, several papers have proposed reinforcement learning policies~\cite{cubuk2019autoaugment,ho2019population} and curriculum learning strategies~\cite{wei2021few,zhang2021flexmatch} to automatically augment data. Also, a couple of consistency regularization methods are introduced to simplify the semi-supervised learning process~\cite{berthelot2019remixmatch,sohn2020fixmatch} as well as to boost performance in domain adaptation scenarios~\cite{berthelot2021adamatch} to improve semi-supervised learning. As far as we know, our work is the first to apply a meta-learner to automatically determine the augmentation strength in consistency regularization in semi-supervised text classification.
\section{Conclusion}
In closing, this paper has proposed an instance-adaptive self-training method SAT to boost performance in semi-supervised text classification. Inspired by FixMatch, SAT combines data augmentation and consistency regularization and designs a novel meta-learner to automatically determine the relative strength of augmentations. Empirical experiments and ablation studies confirm SAT is simple yet effective in improving semi-supervised learning.

\section*{Limitations}
Our proposed method has two limitations. First, in our experiments, we found the semi-supervised process is easy to be influenced by unlabeled data size. During training, we adjusted the size of unlabeled data in each batch by adjusting the $\mu$ value, as mentioned in~\cref{sec:problem_setting}. It will be a future direction that we design some strategies to automatically learn the $\mu$ value. Second, this work only considered the situation where there are only two augmentations in consistency regularization. As our SAT method can automatically rank the augmentation strengths, our future work is to extend SAT to regularize more than two augmentations.

\section*{Ethical Considerations}
Augmentation techniques discussed in~\cref{sec:aug_methods} should be used with care since they might generate data that do not align with the original meaning. 

\section*{Acknowledgments}
We would like to thank the anonymous reviewers for their helpful comments. This research is supported by the Ministry of Education, Singapore, under its AcRF Tier-2 grant (Project no. T2MOE2008, and Grantor reference no. MOET2EP20220-0017). Any opinions, findings, conclusions, or recommendations expressed in this material are those of the authors and do not reflect the views of the Ministry of Education, Singapore. 

\bibliography{anthology,custom}
\bibliographystyle{acl_natbib}

\appendix

\section{Dataset Statistics}
\label{sec:appendix1}
The dataset statistics and split information are presented in~\cref{tab:data_stat}.
\begin{table}[ht]
    \centering
    \begin{tabular}{c|ccc}
    \toprule
         Datasets&\# Unlabeled & \# Dev & \# Test\\
    \midrule
         AG News & 5000 & 2000 & 1900 \\
         Yahoo! & 5000 & 2000 & 6000 \\
         IMDB & 5000 & 1000 & 12500 \\
    \bottomrule
    \end{tabular}
    \caption{Dataset statistics and data splits. The number of unlabeled data, dev data and test data in the table means the number of data per class.}
    \label{tab:data_stat}
\end{table}

\section{Implementation Details}
We performed a grid search for hyperparameters: $\eta_{main} \in \{5e-5, 1e-3\}, \eta_{bert} \in \{1e-5, 5e-5\}$, $\beta$ is fixed at $1e^{-4}$, $\mu \in \{3,4,5,8,10,20\}$, $\tau\in\{0.90,0.95,0.99\}$, and batch size is fixed at 32. We tuned our model on a single NVIDIA RTX 8000 GPU. We ran each experiment for 50 epochs with a patience of 15 or 10 for early stopping. 

\end{document}